\documentclass[11pt,a4paper,UTF8]{article}
\usepackage[hyperref]{emnlp2020}
\usepackage{times}
\usepackage{CJKutf8}
\usepackage{latexsym}
\usepackage{soul}
\usepackage{url}
\usepackage{graphicx}
\usepackage{amsmath}
\usepackage{amsthm}
\usepackage{booktabs}
\usepackage{algorithm}
\usepackage{algorithmic}
\usepackage{amsfonts}
\usepackage{multirow}

\usepackage{microtype}

\aclfinalcopy 

\title{BERT for Monolingual and Cross-Lingual Reverse Dictionary}

\author{Hang Yan, Xiaonan Li, Xipeng Qiu\thanks{\ \  Corresponding author.} ,  Bocao Deng \\
  Shanghai Key Laboratory of Intelligent Information Processing, Fudan University \\
  School of Computer Science, Fudan University \\
  2005 Songhu Road, Shanghai, China \\
  \texttt{\{hyan19,xnli20,xpqiu\}@fudan.edu.cn, dengbocao@gmail.com }}
\date{}

\begin{document}
\maketitle

\begin{abstract}
  Reverse dictionary is the task to find the proper target word given the word description. In this paper, we tried to incorporate BERT into this task. However, since BERT is based on the byte-pair-encoding (BPE) subword encoding, it is nontrivial to make BERT generate a word given the description. We propose a simple but effective method to make BERT generate the target word for this specific task. Besides, the cross-lingual reverse dictionary is the task to find the proper target word described in another language. Previous models have to keep two different word embeddings and learn to align these embeddings. Nevertheless, by using the Multilingual BERT (mBERT), we can efficiently conduct the cross-lingual reverse dictionary with one subword embedding, and the alignment between languages is not necessary. More importantly, mBERT can achieve remarkable cross-lingual reverse dictionary performance even without the parallel corpus, which means it can conduct the cross-lingual reverse dictionary with only corresponding monolingual data. Code is publicly available at \url{https://github.com/yhcc/BertForRD.git}.
\end{abstract}

\section{Introduction}

\begin{figure}[t]
    \centering
    \includegraphics[width=\columnwidth]{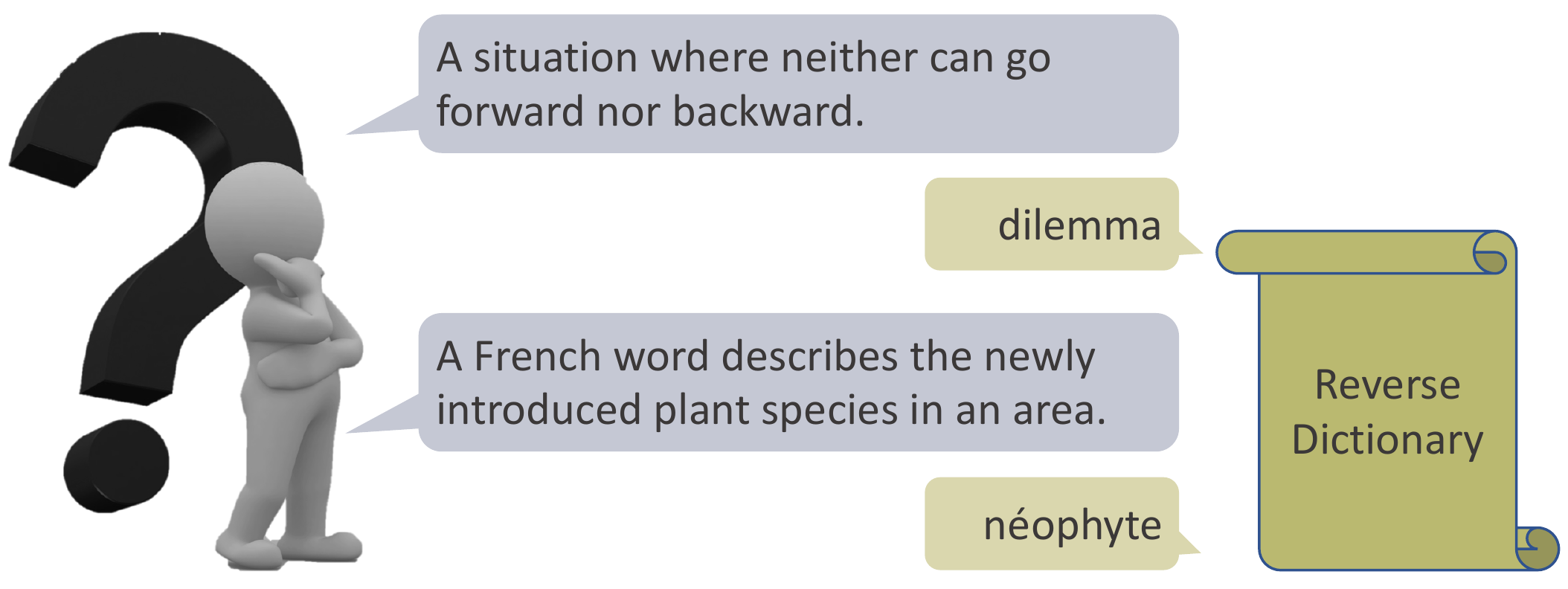}
    \caption{An example of the monolingual and cross-lingual reverse dictionary. }\label{fig:example}
\end{figure}

Reverse dictionary \cite{bilac2004dictionary,Hill2016LearningTU} is the task to find the proper target word given the word description.
Fig. \ref{fig:example} shows an example of the monolingual and the cross-lingual reverse dictionary. Reverse dictionary should be a useful tool to help writers, translators, and new language learners find a proper word when encountering the tip-of-the-tongue problem  \cite{brown1966tip}.
Moreover, the reverse dictionary can be used for educational evaluation. For example, teachers can ask the students to describe a word, and the correct description should make the reverse dictionary model recall the word.


The core of reverse dictionary is to match a word and its description semantically. Early methods \cite{bilac2004dictionary,Shaw2013BuildingAS} firstly extracted the handcrafted features and then used similarity-based approaches to find the target word. However, since these methods are mainly based on the surface form of words, they cannot extract the semantic meaning, resulting in bad performance when evaluated on the human-written search query.
Recent methods usually adopt neural networks to encode the description and the candidate words into the same semantic embedding space and return the word which is closest to the description \cite{Hill2016LearningTU,zhang2019multi}.

Although current neural methods can extract the semantic representations of the descriptions and words, they have three challenging issues: (1) The first issue is the data sparsity.  It is hard to learn good embeddings for the low-frequent words; (2) The second issue is polysemy. The previous methods usually use the static word embedding \cite{Mikolov:2013,pennington2014glove}, making them struggle to find the target word when the target word is polysemous. \citet{pilehvar2019importance} used different word senses to represent a word. Nonetheless, gathering senses for all words is not easy;
(3) The third issue is the alignment of cross-lingual word embeddings in the cross-lingual reverse dictionary scenario \cite{Hill2016LearningTU,DBLP:journals/corr/abs-1808-03726}.

In this paper, we leverage the pre-trained masked language model BERT \cite{DBLP:conf/naacl/DevlinCLT19} to tackle the above issues. Firstly, since BERT tokenizes the words into subwords with byte-pair-encoding (BPE) \cite{DBLP:conf/acl/SennrichHB16a}, the common subwords between low-frequent and high-frequent words can alleviate the data sparsity problem. Secondly, BERT can output contextualized representation for a word. Thus the polysemy problem can be much relieved. Thirdly, the mBERT is suitable to tackle the cross-lingual reverse dictionary. Because BERT shares some subwords between different languages, there is no need to align different languages explicitly.
Therefore, we formulate the reverse dictionary task into the masked language model framework and use BERT to deal with the reverse dictionary task in monolingual and cross-lingual scenarios.
Besides, our proposed framework can also tackle the cross-lingual reverse dictionary task without the parallel (aligned) corpus.  

Our contributions can be summarized as follows:
\begin{enumerate}
 \item We propose a simple but effective solution to incorporate BERT into the reverse dictionary task. In the method, the target word is predicted according to masked language model predictions. With BERT, we achieve significant improvement for the monolingual reverse dictionary task.
 \item By leveraging the Multilingual BERT (mBERT), we extend our methods into the cross-lingual reverse dictionary task, mBERT can not only avoid the explicit alignment between different language embeddings, but also achieve good performance.
 \item We propose the unaligned cross-lingual reverse dictionary scenario and achieve encouraging performance only with monolingual reverse dictionary data. As far as we know, this is the first time the unaligned cross-lingual reverse dictionary is inspected.
\end{enumerate}


\section{Related Work}
The reverse dictionary task has been investigated in several previous academic studies. \citet{bilac2004dictionary} proposed using the information retrieval techniques to solve this task, and they first built a database based on available dictionaries. When a query came in, the system would find the closest definition in the database, then return the corresponding word. Different similarity metrics can be used to calculate the distance. \citet{Shaw2013BuildingAS} enhanced the retrieval system with WordNet \cite{Miller1995WordNet}. \citet{Hill2016LearningTU} was the first to apply RNN into the reverse dictionary task, making the model free of handcrafted features. After encoding the definition into a dense vector, this vector is used to find its nearest neighbor word. This model formulation has been adopted in several papers \cite{pilehvar2019importance,DBLP:journals/corr/abs-1808-03726,zhang2019multi,MorinagaY18,hedderich2019using}, their difference lies in usage of different resources. \citet{kartsaklis2018mapping,Thorat2016ImplementingAR} used WordNet to form graphs to tackle the reverse dictionary task.

The construction of the bilingual reverse dictionary has been studied in \cite{DBLP:conf/sigir/GollinsS01,Lam2013CreatingRB}. \citet{Lam2013CreatingRB} relied on the availability of lexical resources, such as WordNet, to build a bilingual reverse dictionary. \citet{DBLP:journals/corr/abs-1808-03726} built several bilingual reverse dictionaries based on the Wiktionary\footnote{\url{https://www.wiktionary.org/}}, but this kind of online data cannot ensure the data's quality. Building a bilingual reverse dictionary is not an easy task, and it will be even harder for low-resource language. Other than the low-quality problem, the vast number of language pairs is also a big obstacle, since if there are $N$ languages, they will form $N^2$ pairs. However, by the unaligned cross-lingual reverse dictionary, we can not only exploit the high-quality monolingual dictionaries, but also avoid the preparation of $N^2$ language pairs.

Unsupervised machine translation is highly correlated with the unaligned cross-lingual reverse dictionary \cite{DBLP:conf/iclr/LampleCDR18,DBLP:conf/nips/ConneauL19,DBLP:conf/acl/SennrichHB16}. However, the unaligned cross-lingual reverse dictionary task differs from the unsupervised machine translation at least in two aspects. Firstly, the target for the cross-lingual reverse dictionary and machine translation is a word and a sentence, respectively. 
Secondly, theoretically, the translated sentence and the original sentence should contain the same information.
Nevertheless, in the cross-lingual reverse dictionary task, on the one hand, the target word might contain more senses when it is polysemous. On the other hand, a description can correspond to several similar terms. The polysemy also makes the unsupervised word alignment hard to solve this task \cite{DBLP:conf/iclr/LampleCRDJ18}.

Last but not least, the pre-trained language model BERT has been extensively exploited in the Natural Language Processing (NLP) community since its introduction \cite{DBLP:conf/naacl/DevlinCLT19,DBLP:conf/nips/ConneauL19}. Owing to BERT's ability to extract contextualized information, BERT has been successfully utilized to enhance various tasks substantially, such as the aspect-based sentiment analysis task \cite{DBLP:conf/naacl/SunHQ19}, summarization \cite{DBLP:conf/acl/ZhongLWQH19}, named entity recognition \cite{DBLP:journals/corr/abs-1911-04474,DBLP:conf/acl/LiYQH20} and Chinese dependency parsing \cite{DBLP:journals/corr/abs-1904-04697}. However, most works used BERT as an encoder, and less work uses BERT to do generation \cite{DBLP:journals/corr/abs-1902-04094,DBLP:conf/nips/ConneauL19}. \citet{DBLP:journals/corr/abs-1902-04094} showed that BERT is a Markov random field language model. Therefore, sentences can be sampled from BERT. \citet{DBLP:conf/nips/ConneauL19} used pre-trained BERT to initialize the unsupervised machine training model an achieve good performance. Different from these work, although a word might contain several subwords, we use a simple but effective method to make BERT generate the word ranking list with only one forward pass.

\section{Methodology}

The reverse dictionary task is to find the target word $w$ given its definition $d=[w_1, w_2, \ldots, w_n]$, where $d$ and $w$ can be in the same language or different languages. In this section, we first introduce BERT, then present the method we used to incorporate BERT into the reverse dictionary task.

\subsection{BERT}
BERT is a pre-trained  model proposed in \cite{DBLP:conf/naacl/DevlinCLT19}. BERT contains several Transformer Encoder layers. BERT can be formulated as follows
\begin{align}
     & \hat{h^l} = \mathrm{LN}(h^{l-1} + \mathrm{MHAtt}(h^{l-1})),\\
     & h^l = \mathrm{LN}(\hat{h^l} + \mathrm{FFN}(\hat{h^l})),
\end{align}
where $h^0$ is the BERT input, for each token, it is the sum of its token embedding, position embedding, and segment embedding; $\mathrm{LN}$ is the layer normalization layer; $\mathrm{MHAtt}$ is the multi-head self-attention; $\mathrm{FFN}$ contains three layers, the first one is a linear projection layer, then an activation layer, then another linear projection layer; $l$ is the depth of the layer, the total number of layers in BERT is 12 or 24.

Two tasks were used to pre-train BERT. The first is to replace some tokens with the ``[MASK]'' symbol, BERT has to recover this masked token from outputs of the last layer. The second one is the next sentence prediction. For two continuous sentences, 50\% of the time the second sentence will be replaced with other sentences, BERT has to figure out whether the input sequence is continuous based on the output vector of the ``[CLS]'' token. Another noticeable fact about BERT is that, instead of directly using the word, it used BPE subword \cite{DBLP:conf/acl/SennrichHB16a} to represent tokens.Therefore, one word may be split into several tokens. Next, we will show how we make BERT generate the word ranking list.

\begin{figure}[]
    \centering
    \includegraphics[width=\columnwidth]{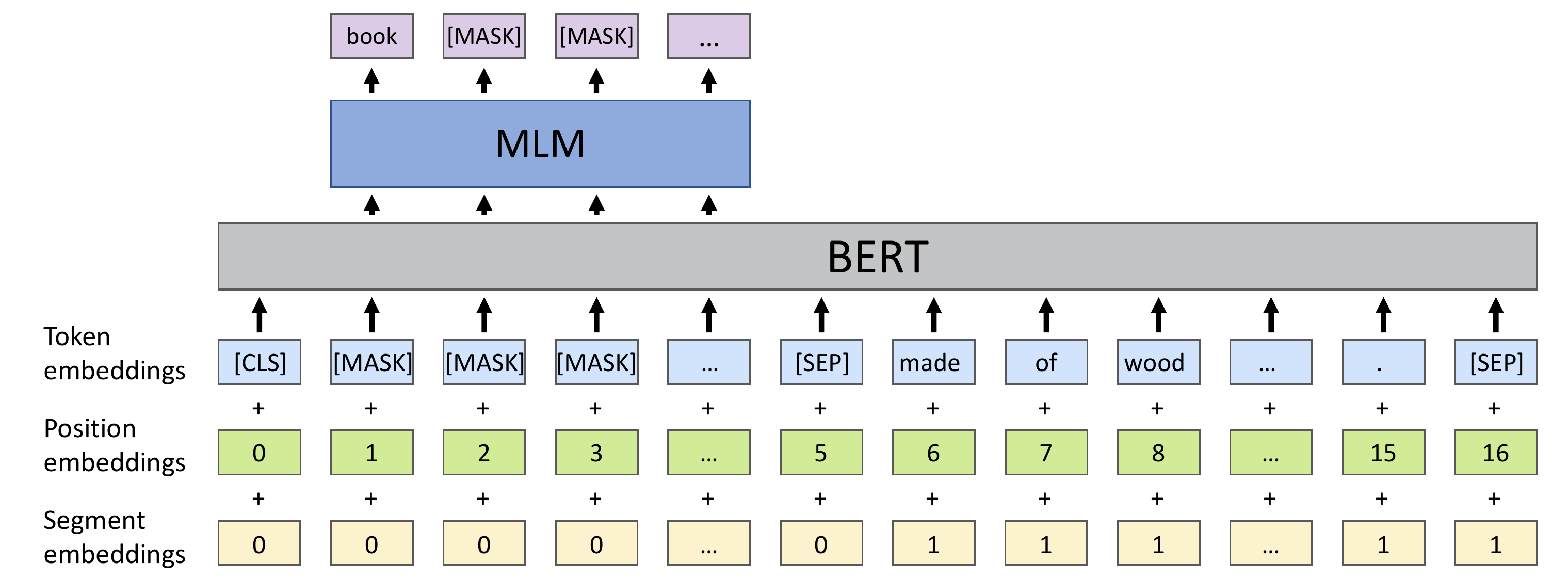}
    \caption{The model structure for the monolingual and cross-lingual reverse dictionary. The ``[MASK]'' in the input is the placeholder where BERT needs to predict. Placeholders concatenate with the word definition before sending it into BERT. Postprocessing is required to convert the prediction for ``[MASK]''s into the word ranking list. }\label{fig:model1}
\end{figure}

\subsection{BERT for Monolingual Reverse Dictionary}
The model structure is shown in Fig. \ref{fig:model1}. The input sequence $x$ has the form ``[CLS] + [MASK] * $k$ + [SEP] + [subword sequence of the definition $d$] + [SEP]''. We want BERT to recover the target word $w$ from the $k$ ``[MASK]'' tokens based on the definition $d$. 
We first utilize BERT to predict the masks as in its pre-training task. It can be formulated as 
\begin{align}
    S_{subword} = \mathrm{MLM}(H^{L}_k),
\end{align}
where  $H^L_k \in \mathbb{R}^{k \times d_{model}}$ is the hidden states for the $k$ masked tokens in the last layer, $\mathrm{MLM}$ is the pre-trained masked language model, $S_{subword} \in \mathbb{R}^{k \times |V|}$ is the subword score distribution for the $k$ positions, $|V|$ is the number of subword tokens.  
Although we can make BERT directly predict word by using a word embedding, it will suffer from at least two problems: the first one is that it cannot take advantage of common subwords between words, such as prefixes and postfixes; the second one is that predicting word is inconsistent with the pre-trained tasks.

After achieving $S_{subword}$, we need to convert them back to word scores. However, there are $|V|^k$ kinds of subword combinations, which makes it intractable to represent words by crossing subwords. Another method is to generate subword one-by-one \cite{DBLP:journals/corr/abs-1902-04094,DBLP:conf/nips/ConneauL19}, it is not suitable for this task, since this task needs to return a ranking list of words, but the generation can only offer limited answers. Nevertheless, for this specific task, the number of possible target words is fixed since the number of unique words in one language's dictionary is limited. Hence, instead of combining the subword sequence into different words, we can only care for the subword sequence, which can form a valid word.

Specifically, for a given language, we first list all its valid words and find the subword sequence for each word. For a word $w$ with the subword sequence $[b_1,...,b_k]$, its score is calculated by 
\begin{align}
    S_{word} = \sum_{i=1}^k S_{subword}^i[b_i], \label{eq:subword2word}
\end{align}
where $S_{word} \in \mathbb{R}$ is the score for the word $w$, $S_{subword}^i \in \mathbb{R}^{|V|}$ is the subword score distribution in the $i$th position, $S_{subword}^i [b_i]$ is gathering the $b_i$th element in $S_{subword}^i$. However, not all words can be decomposed to $k$ subword tokens. If a word has subword tokens less than $k$, we pad it with ``[MASK]'', while our method cannot handle words with more than $k$ subword tokens. By this method, each word can get a score. Therefore we can directly use the cross-entropy loss to finetune the model,
\begin{align}
  L_w = - \sum_{i=1}^N w^{(i)}\mathrm{log\_softmax}(S_{word}^{(i)}),
\end{align}
where $N$ is the total number of samples, $w$ is the target word.
When ranking, words are sorted by their scores.

\subsection{BERT for Cross-lingual Reverse Dictionary}
The model structure used in this setting is as depicted in Fig. \ref{fig:model1}. The only difference between this setting and the monolingual scenario is the pre-trained model used. This setting uses the mBERT model. mBERT has the same structure as BERT, but it was trained on 104 languages. Therefore its token embedding contains subwords in different languages.

\subsection{BERT for Unaligned Cross-lingual Reverse Dictionary}
\begin{figure}[]
  \centering
  \includegraphics[width=\columnwidth]{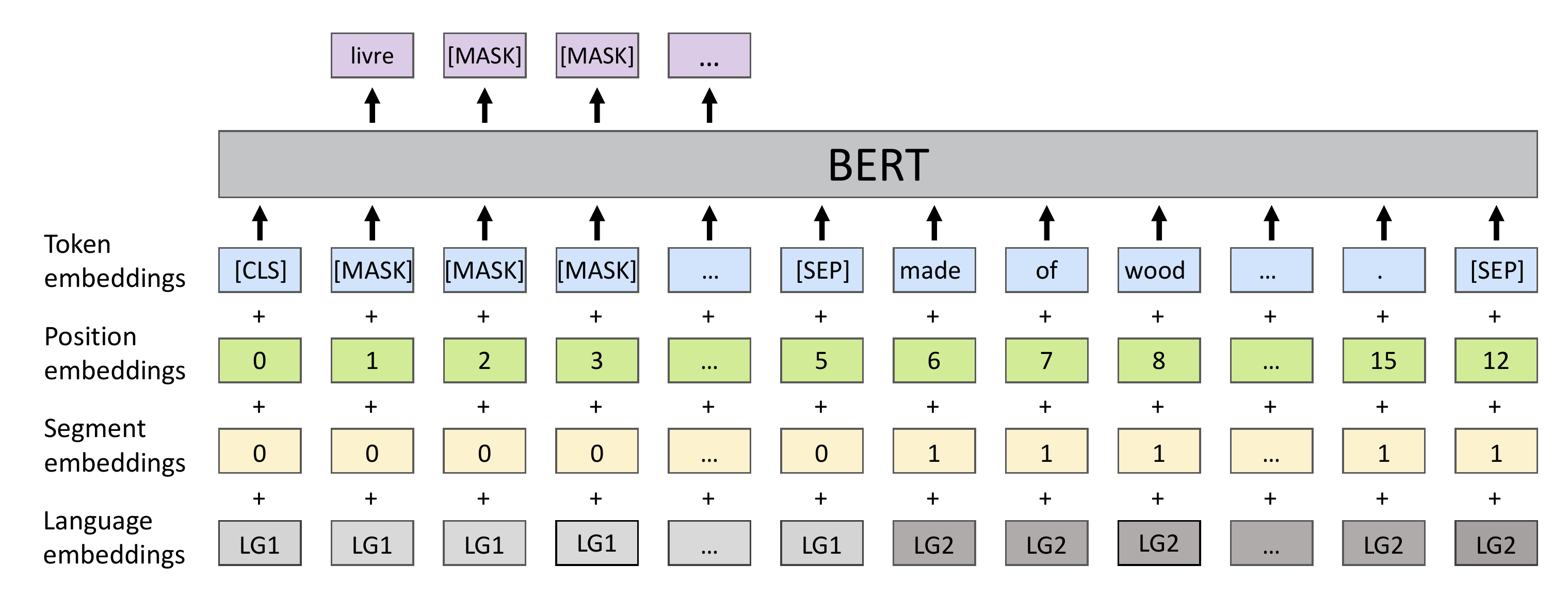}
  \caption{The model structure for the unaligned cross-lingual reverse dictionary. We add a randomly initialized language embedding to distinguish languages. Since we only have monolingual training data, ``LG1'' and ``LG2'' are of the same value in the training phase, but different in the evaluation phase.}\label{fig:model2}
\end{figure}

The model used for this setting is as depicted in Fig. \ref{fig:model2}. 
Compared with the BERT model, we add an extra learnable language embedding in the bottom, and the language embedding has the same dimension as the other embeddings. Except for the randomly initialized language embedding, the model is initialized with the pre-trained mBERT. 

Instead of using the $\mathrm{MLM}$ to get $S_{subword}$, we use the following equation to get $S_{subword}$

\begin{align}
    S_{subword} = H^{L}_k Emb_{token}^T, \label{eq:sbpe}
\end{align}
where $Emb_{token} \in \mathbb{R}^{ |V| \times d_{model}}$ is the subword token embeddings. We found this formulation will lead to better performance than using the $\mathrm{MLM}$, and we assume this is because the training data only contains monolingual data, thus it will be hard for the model to predict tokens in another language when evaluation, while if the $Emb_{token}$ is used, the model can utilize the similarity between subwords to make reasonable predictions. After getting $S_{subword}$, we use Eq.\ref{eq:subword2word} to get the scores for each word, and different languages have different word lists, the loss is calculated by 

\begin{align}
  L_w = - \sum_{j=1}^M \sum_{i=1}^{N_j} w_j^{(i)} \mathrm{log\_softmax}(S_{word_j}^{(i)}),
\end{align}
where $M$ is the number languages, $N_k$ is the number of samples for language $j$, $w_j^{(i)}$ is the target word in language $j$, $S^{(i)}_{word_j}$ is the score distribution for words in language $j$. When getting the ranking list for a language, we only calculate word scores for that language.

\begin{table}\scriptsize
    \addtolength{\tabcolsep}{-4.5pt}
    \begin{tabular}{ccccccccc}
    \toprule
    Language                 & Word          & Type & Train  & Dev   & Seen & Unseen & Description & Question \\ \midrule
    \multirow{2}{*}{English} & \multirow{2}{*}{50.5K} & Def & 675.7K & 75.9K & 500  & 500    & 200         & -        \\
                             &                        & Word & 45.0K  & 5.0K  & 500  & 500    & 200         & -        \\ \midrule
    \multirow{2}{*}{Chinese} & \multirow{2}{*}{58.5K} & Def & 78.3K  & 8.7K  & 2.1K & 2.0K   & 200         & 272      \\
                             &                        & Word & 54.0K  & 6.1K  & 1.4K & 1.4K   & 200         & 272      \\ \bottomrule
    \end{tabular}
    \caption{Dataset statistics for the monolingual reverse dictionary. The row ``Def'' and ``Word'' are the number of definition and distinct words in the split, respectively.} \label{tb:stat1}
\end{table}

\section{Experimental Setup}


\subsection{Dataset} \label{sec:dataset}
For the monolingual reverse dictionary, we tested our methods in the English dataset and Chinese dataset released by \cite{Hill2016LearningTU} and \cite{zhang2019multi}, respectively. \citet{Hill2016LearningTU} built this dataset by extracting words and definitions from five electronic dictionaries and Wikipedia. \citet{zhang2019multi} used the authoritative Modern Chinese Dictionary to build the Chinese reverse dictionary. There are four different test sets: (1) \textbf{Seen} definition set, words and their definitions are seen during the training phase; (2) \textbf{Unseen} definition set, none of the word's definitions have been seen during the training phase, but they might occur in other words' definition; (3) \textbf{Description} definition set, the description and its corresponding word are given by human. 
Methods rely on word matching may not perform well in this setting \cite{Hill2016LearningTU}; (4) \textbf{Question} definition set, this dataset is only in Chinese, it contains 272 definitions appeared in Chinese exams. The detailed dataset statistics are shown in Table \ref{tb:stat1}.

For the cross-lingual and unaligned cross-lingual reverse dictionary, we use the dataset released in \cite{DBLP:journals/corr/abs-1808-03726}. This dataset includes four bilingual reverse dictionaries: English$\leftrightarrow$French, English$\leftrightarrow$Spanish. Besides, this dataset includes English, French, and Spanish monolingual reverse dictionary data. The test set for this dataset is four bilingual reverse dictionaries: En$\leftrightarrow$Fr and En$\leftrightarrow$Es. For the cross-lingual reverse dictionary, we use the paired bilingual reverse dictionary data to train our model; for the unaligned cross-lingual reverse dictionary, we use the three monolingual reverse dictionary data to train our model. And for both settings, we report results on the test sets of the four bilingual reverse dictionary. The detailed dataset statistics are shown in Table \ref{tb:stat2}. 


\begin{table}[]\scriptsize
    \addtolength{\tabcolsep}{-0.4pt}
    \begin{tabular}{ccccccc}
        \toprule
        Scenario                     & Language               & Word                    & Type & Train  & Dev & Test \\ \midrule
        \multirow{6}{*}{Monolingual} & \multirow{2}{*}{En}    & \multirow{2}{*}{117.4K} & Def & 228.2K & 500 & 501  \\
                                        &                        &                         & Word & 117.3K & 499 & 501  \\ \cmidrule{2-7}
                                        & \multirow{2}{*}{Fr}    & \multirow{2}{*}{52.4K}  & Def & 104.4K & 500 & 501  \\
                                        &                        &                         & Word & 52.2K  & 496 & 501  \\ \cmidrule{2-7}
                                        & \multirow{2}{*}{Es}    & \multirow{2}{*}{22.5K}  & Def & 47.6K  & 500 & 501  \\
                                        &                        &                         & Word & 22.4K  & 493 & 501  \\ \midrule
        \multirow{8}{*}{Bilingual}   & \multirow{2}{*}{En-Fr} & \multirow{2}{*}{45.6K}  & Def & 49.7K  & 500 & 501  \\
                                        &                        &                         & Word & 15.6K  & 493 & 488  \\ \cmidrule{2-7}
                                        & \multirow{2}{*}{Fr-En} & \multirow{2}{*}{44.5K}  & Def & 58.1K  & 500 & 501  \\
                                        &                        &                         & Word & 16.8K  & 487 & 486  \\ \cmidrule{2-7}
                                        & \multirow{2}{*}{En-Es} & \multirow{2}{*}{45.6K}  & Def & 20.2K  & 500 & 501  \\
                                        &                        &                         & Word & 7.9K   & 484 & 495  \\ \cmidrule{2-7}
                                        & \multirow{2}{*}{Es-En} & \multirow{2}{*}{35.8K}  & Def & 55.9K  & 500 & 501  \\
                                        &                        &                         & Word & 15.9K  & 489 & 487  \\ \bottomrule
    \end{tabular}
    \caption{Dataset statistics for the cross-lingual and unaligned cross-lingual reverse dictionary. The upper block is the monolingual data used to train the unaligned cross-lingual reverse dictionary. The lower block is the cross-lingual reverse dictionary data. Both scenarios were evaluated in the test set in the lower part. ``En-fr'' means the target word is in English, the definition is in French.} \label{tb:stat2}
\end{table}

\subsection{Evaluation Metrics}
For the English and Chinese monolingual reverse dictionary, we report three metrics: the median rank of target words (Median Rank, lower better, lowerest is 0), the ratio that target words appear in top 1/10/100 (Acc@1/10/100, higher better, ranges from 0 to 1), and the variance of the rank of the correct target word (Rank Variance, lower better), these results are also reported in \cite{Hill2016LearningTU,zhang2019multi}. For the cross-lingual and unaligned cross-lingual reverse dictionary, we report the Acc@1/10, and the mean reciprocal rank (MRR, higher is better, ranges from 0 to 1), these results are also reported in \cite{DBLP:journals/corr/abs-1808-03726}.

\subsection{Hyper-parameter Settings} \label{sec:hyper}
The English BERT and Multilingual BERT (mBERT) are from \cite{DBLP:conf/naacl/DevlinCLT19}, the Chinese BERT is from \cite{DBLP:journals/corr/abs-1906-08101}. Since RoBERTa has the same model structure as BERT, we also report the performance with the English RoBERTa from \cite{DBLP:journals/corr/abs-1907-11692} and the Chinese RoBERTa from \cite{DBLP:journals/corr/abs-1906-08101} for the monolingual reverse dictionary. Both RoBERTa and BERT are the base version, and we use the uncased English BERT and cased mBERT. For all models, we find the hyper-parameters based on the Acc@10 in the development sets, the models with the best development set performance are evaluated on the test set. The data and detailed hyper-parameters for each setting will be released within the code \footnote{\url{https://github.com/yhcc/BertForRD.git}}. 
 We choose $k=4$ for Chinese, and $k=5$ for other languages, $k$ is determined by at least 99\% of the target words in the training set are included. 

 \section{Experimental Results}

 \subsection{Monolingual Reverse Dictionary}  \label{sec:mono}

 Results for the English and Chinese monolingual reverse dictionary have been shown in Table \ref{tb:en} and Table \ref{tb:cn}, respectively. ``OneLook'' in Table \ref{tb:en} is the most used commercial reverse dictionary system, it indexed over 1061 dictionaries, even included online dictionaries, such as Wikipedia and WordNet \cite{Miller1995WordNet}. Therefore, its result in the unseen definition test set is ignored. ``SuperSense'', ``RDWECI'', ``MS-LSTM'' and ``Mul-Channel'' are from \cite{pilehvar2019importance,MorinagaY18,kartsaklis2018mapping,zhang2019multi}, respectively. From Table \ref{tb:en}, RoBERTa achieves state-of-the-art performance on the human description test set. And owing to bigger models, in the seen definition test set, compared with the ``Mul-channel'', BERT and RoBERTa enhance the performance significantly. Although the MS-LSTM \cite{kartsaklis2018mapping} performs remarkably in the seen test sets, it fails to generalize to unseen and description test sets. Besides, ``RDWECI'', ``SuperSense'', ``Mul-channel'' in Table \ref{tb:en} all used external knowledge, such as WordNet, Part-of-Speech tags. Combining BERT and structured knowledge should further improve the performance in all test sets, we leave it for further work.
 
 Table \ref{tb:cn} presents the results for the Chinese reverse dictionary. 
For the seen definition setting, BERT and RoBERTa substantially improve the performance. 
 Apart from the good performance in seen definitions, BERT and RoBERTa perform well in the human description test set, which depicts their capability to capture human's meaning.

\begin{table}[h]\scriptsize
    \addtolength{\tabcolsep}{-4.6pt}
        \begin{tabular}{cccccccccc}
            \toprule
            \multicolumn{1}{c|}{Model}         & \multicolumn{3}{c|}{Seen}     & \multicolumn{3}{c|}{Unseen}                                                      & \multicolumn{3}{c}{Description}                                                                    \\ \midrule
            \multicolumn{1}{c|}{OneLook*}    & 0                              & .66/.94/.95                            & \multicolumn{1}{c|}{200} & -                       & -                                      & \multicolumn{1}{c|}{-}   & 5.5                            & \textbf{.33}/.54/.76                     & 332                     \\ \midrule
            \multicolumn{1}{c|}{RDWECI}        & 121                            & .06/.20/.44                            & \multicolumn{1}{c|}{420} & 170                     & .05/.19/.43                            & \multicolumn{1}{c|}{420} & 16                             & .14/.41/.74                              & 306                     \\
            \multicolumn{1}{c|}{SuperSense}    & 378                            & .03/.15/.36                            & \multicolumn{1}{c|}{462} & 465                     & .02/.11/.31                            & \multicolumn{1}{c|}{454} & 115                            & .03/.15/.47                              & 396                     \\
            \multicolumn{1}{c|}{MS-LSTM*}       & \textbf{0}                     & \textbf{.92}/\textbf{.98}/\textbf{.99} & \multicolumn{1}{c|}{65}  & 276                     & .03/.14/.37                            & \multicolumn{1}{c|}{426} & 1000                           & .01/.04/.18                              & 404                     \\
            \multicolumn{1}{c|}{Mul-Channel} & 16                             & .20/.44/.71                            & \multicolumn{1}{c|}{310} & 54                      & .09/.29/.58 & \multicolumn{1}{c|}{358} & 2                              & .32/.64/.88                              & 203                     \\ \midrule
            \multicolumn{1}{c|}{BERT}           & \textbf{0} & .57/.86/.92     & \multicolumn{1}{c|}{240}  & \textbf{18} & \textbf{.20/.46/.64}       & \multicolumn{1}{c|}{418}  & 1 & .36/.77/.94 & 94 \\ 

            \multicolumn{1}{c|}{RoBERTa}           & \textbf{0} & .57/.84/.92     & \multicolumn{1}{c|}{228}  & 37 & .10/.36/.60       & \multicolumn{1}{c|}{405}  & \textbf{1} & \textbf{.43/.85/.96} & 46 \\ \bottomrule
        \end{tabular}
    \caption{Results on the English reverse dictionary datasets. In each cell, the values are the ``Median Rank'', ``Acc@1/10/100'' and ``Rank Variance''. * results are from \protect\cite{zhang2019multi} . BERT and RoBERTa achieve a significant performance boost in both the description test set and the unseen test set.} \label{tb:en}
\end{table}

\begin{table}[]\scriptsize
    \centering
    \addtolength{\tabcolsep}{-4.7pt}
        \begin{tabular}{ccccccccccccc}
            \toprule
            \multicolumn{1}{c|}{Model}         & \multicolumn{3}{c|}{Seen}                                                                  & \multicolumn{3}{c|}{Unseen}  & \multicolumn{3}{c|}{Description}  & \multicolumn{3}{c}{Question}                                                                       \\ \midrule
            \multicolumn{1}{c|}{BOW*}           & 59                             & .08/.28                               & \multicolumn{1}{c|}{403} & 65                     & .08/.28                              & \multicolumn{1}{c|}{411}         & 40                             & .07/.30                              & \multicolumn{1}{c|}{357} & 42                             & .10/.28                              & 362                     \\
            \multicolumn{1}{c|}{RDWECI*}        & 56                             & .09/.31                               & \multicolumn{1}{c|}{423} & 83                     & .08/.28                              & \multicolumn{1}{c|}{436}         & 32                             & .09/.32                              & \multicolumn{1}{c|}{376} & 45                             & .12/.32                              & 384                     \\
            \multicolumn{1}{c|}{Mul-Channel*} & 1                              & .49/.78                               & \multicolumn{1}{c|}{220} & 10                     & .18/.49            & \multicolumn{1}{c|}{310}         & 5                              & .24/.56                              & \multicolumn{1}{c|}{260} & 0                              & .50/.73            & 223            \\ \midrule
            \multicolumn{1}{c|}{BERT}           & \textbf{0} & \textbf{.88/.93} & \multicolumn{1}{c|}{201}  & \textbf{5} & \textbf{.27/.56} & \multicolumn{1}{c|}{360}     & \textbf{3} & \textbf{.34/.67} & \multicolumn{1}{c|}{260}  & \textbf{0} & \textbf{.57}/.70 & \multicolumn{1}{c}{325} \\

            \multicolumn{1}{c|}{RoBERTa}           & \textbf{0} & \textbf{.88/.93} & \multicolumn{1}{c|}{200}  & \textbf{5} & \textbf{.28/.56} & \multicolumn{1}{c|}{350}     & \textbf{3} & \textbf{.33/.65} & \multicolumn{1}{c|}{230}  & \textbf{0} & \textbf{.59/.74} & \multicolumn{1}{c}{310} \\
            \bottomrule
        \end{tabular}
    \caption{Results on the Chinese reverse dictionary datasets. In each cell, the values are the ``Median Rank'', ``Acc@1/10'' and ``Rank Variance''. * results are from \protect\cite{zhang2019multi}. Our proposed methods enhance the performance in all test sets substantially.\label{tb:cn}}
 \end{table}

\subsection{Cross-lingual Reverse Dictionary}

In this section, we will present the results for the cross-lingual reverse dictionary. 
The performance comparison is shown in Table \ref{tb:multi}, mBERT substantially enhances the performance in four test sets. The contrast between ``mBERT'' and ``mBERT-joint'' shows that jointly train the reverse dictionary in different language pairs can improve the performance. 

\begin{table}[]\scriptsize
    \addtolength{\tabcolsep}{-2pt}
        \begin{tabular}{clrlr|lrlr}
            \toprule
            \multicolumn{1}{c|}{Model}      & \multicolumn{2}{c|}{En-Fr}                  & \multicolumn{2}{c|}{Fr-En}               & \multicolumn{2}{c|}{En-Es}                  & \multicolumn{2}{c}{Es-En} \\ \midrule
            \multicolumn{1}{c|}{ATT*}       & .39/.47          & \multicolumn{1}{r|}{.41} & .40/.50          & .43                   & .52/.59          & \multicolumn{1}{r|}{.53} & .60/.68            & .63   \\
            \multicolumn{1}{c|}{mBERT}       & .88/.90          & \multicolumn{1}{r|}{.89} & .88/.90          & .89                   & .79/.81          & \multicolumn{1}{r|}{.80} & .88/.90            & .89   \\ \midrule
            \multicolumn{1}{c|}{ATT-joint*} & .64/.69          & \multicolumn{1}{r|}{.65} & .68/.75          & .71                   & .69/.73          & \multicolumn{1}{r|}{.70} & .79/.83            & .80   \\
            \multicolumn{1}{c|}{mBERT-joint}   & \textbf{.90/.94} & \multicolumn{1}{r|}{.92} & \textbf{.90/.93} & .91                   & \textbf{.83/.88} & \multicolumn{1}{r|}{.85} & \textbf{.93/.95}   & .93   \\ \bottomrule
        \end{tabular}
    \caption{Results for the cross-lingual reverse dictionary. In each cell, the values are ``Acc@1/10'' and ``MRR''. * results are from \protect\cite{DBLP:journals/corr/abs-1808-03726}. ``En-Fr'' means the target word is in English, while the description is in French. The ``ATT'' and ``mBERT'' used the bilingual corpus to train the model.  The ``ATT-joint'' and ``mBERT-joint'' are trained on four bilingual reverse dictionary corpus simultaneously. }     \label{tb:multi}
\end{table}

\subsection{Unaligned Cross-lingual Reverse Dictionary}
In this section, we present the results of the unaligned bilingual and cross-lingual reverse dictionary. Models are trained on several monolingual reverse dictionary data, but they will be evaluated on bilingual reverse dictionary data. Take the ``En-Fr'' as an example, models are trained on English definitions to English words, French definitions to French words, while in the evaluation phase, the model is asked to recall an English word given the French description or vice versa.

Since previous models do not consider this setting, we make a baseline by firstly getting words with the same language as the definition through a monolingual reverse dictionary model, then using the word translation or aligned word vectors to recall words in another language. Take ``En-Fr'' for instance, we first recall the top 10 French words with the French definition, then each French word is translated into an English word by either translations or word vectors.

\begin{table}[t]
  \scriptsize
  \addtolength{\tabcolsep}{-3.5pt}
      \begin{tabular}{clrlrlr|lr}
          \toprule
          \multicolumn{1}{c|}{Model}                         & \multicolumn{2}{c|}{En-Fr}                      & \multicolumn{2}{c|}{Fr-En}                      & \multicolumn{2}{c|}{En-Es}                   & \multicolumn{2}{c}{Es-En}           \\ \midrule
          \multicolumn{1}{c|}{ATT-joint*} & .64/.69          & \multicolumn{1}{r|}{.65} & .68/.75          & .71                   & .69/.73          & \multicolumn{1}{r|}{.70} & .79/.83            & .80   \\
          \multicolumn{1}{c|}{BERT-joint}   & .90/.94 & \multicolumn{1}{r|}{.92} & .90/.93 & .91                   & .83/.88 & \multicolumn{1}{r|}{.85} & .93/.95   & .93   \\ \midrule
          \multicolumn{1}{c|}{BERT-Match}                    & .35/.41              & \multicolumn{1}{r|}{-}   & .20/.25              & \multicolumn{1}{r|}{-}   & .23/.26              & -                     & .17/.21                       & -   \\
          \multicolumn{1}{c|}{BERT-Trans}                    & .46/.55             & \multicolumn{1}{r|}{-}   & .42/.51              & \multicolumn{1}{r|}{-}   & .44/.49               & -                     & .29/.38                      & -   \\
          \multicolumn{1}{c|}{BERT-Unaligned}                & .70/.\textbf{80}     & \multicolumn{1}{r|}{.74} & .55/.66     & \multicolumn{1}{r|}{.59} & .52/\textbf{.68}     & .58                   & \textbf{.41/.59}              & .48 \\
          \multicolumn{1}{c|}{BERT-joint-Unaligned}            & \textbf{.71/.80}     & \multicolumn{1}{r|}{.74} & \textbf{.56/.67}     & \multicolumn{1}{r|}{.60} & .\textbf{54/.68}     & .59                   & \textbf{.41/.59}              & .47 \\ \bottomrule
      \end{tabular}
  \caption{Results for the unaligned cross-lingual reverse dictionary. In each cell, the values are ``Acc@1/10'' and ``MRR''. * is from \protect\cite{DBLP:journals/corr/abs-1808-03726}. ``En-Fr'' means the target word is in English, while the definition is in French. Models in the lower block do not use aligned data. While models in the upper block use aligned data to train the model. } \label{tb:unalign}
\end{table}

Models listed in Table \ref{tb:unalign} are as follows: (1) \textbf{mBERT-Match} uses aligned word vectors \cite{DBLP:conf/iclr/LampleCRDJ18} to recall the target words in another language; (2) \textbf{mBERT-Trans} uses the translation API\footnote{\url{fanyi.baidu.com}}; (3) \textbf{mBERT-Unaligned} uses two monolingual reverse dictionary corpus to train one model. Therefore, the results of ``En-Fr'' and ``Fr-En'' in Table \ref{tb:unalign} are from the same model; (4) \textbf{mBERT-joint-Unaligned} is trained on all monolingual corpus.

As shown in the Table \ref{tb:unalign}, the ``mBERT-Unaligned'' and ``mBERT-joint-Unaligned'' perform much better than the ``mBERT-Match'' and ``mBERT-Trans''. Therefore, it is meaningful to explore the unaligned reverse dictionary scenario. As we will show in Section \ref{sec:case_study}, the translation method might fail to recall the target words when the word is polysemous.

From Table \ref{tb:unalign}, we can see that jointly training three monolingual reverse dictionary tasks do not help to recall cross-lingual words. Therefore, how to utilize different languages to enhance the performance of the unaligned reverse dictionary is an unsolved problem. Besides, compared with the top block of Table \ref{tb:unalign}, the performance of the unaligned models lags much behind. Hence, there is a lot of room for unaligned performance improvement.

\section{Analysis}

\subsection{Performance for Number of Senses }
\begin{figure}[h]
  \centering
  \includegraphics[width=0.9\columnwidth]{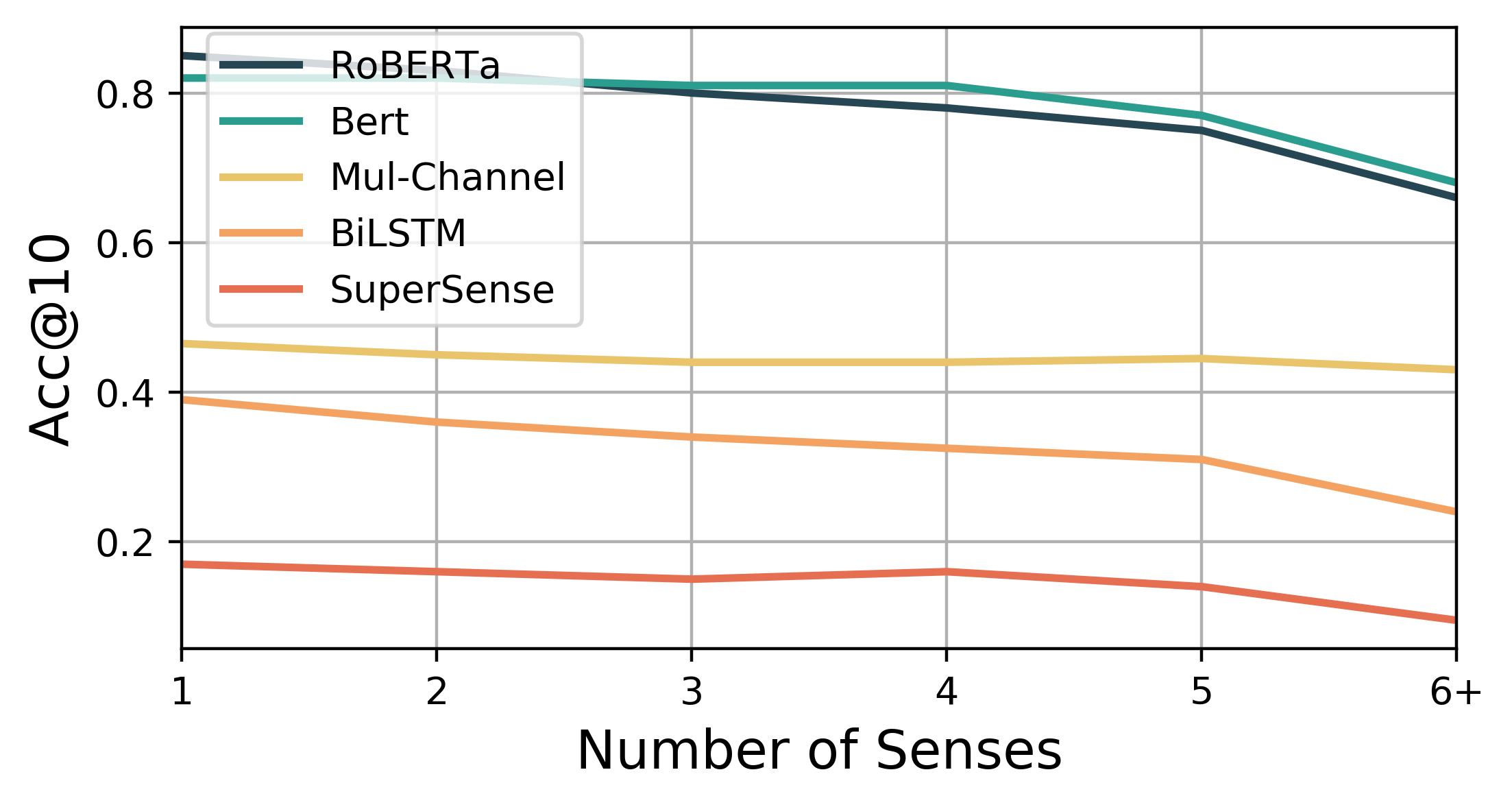}
  \caption{The Acc@10 for English words with a different number of senses.}\label{fig:sense_acc}
\end{figure}

Following \cite{zhang2019multi}, we evaluate the accuracy of words with a different number of senses through WordNet\cite{Miller1995WordNet}. The results are shown in Fig. \ref{fig:sense_acc}. BERT and RoBERTa significantly improve the accuracy of words with single and multiple senses, which means they can alleviate the polysemous issue.

\subsection{Performance for Different Number of Subword}
\begin{figure}[t]
  \centering
  \includegraphics[width=0.9\columnwidth]{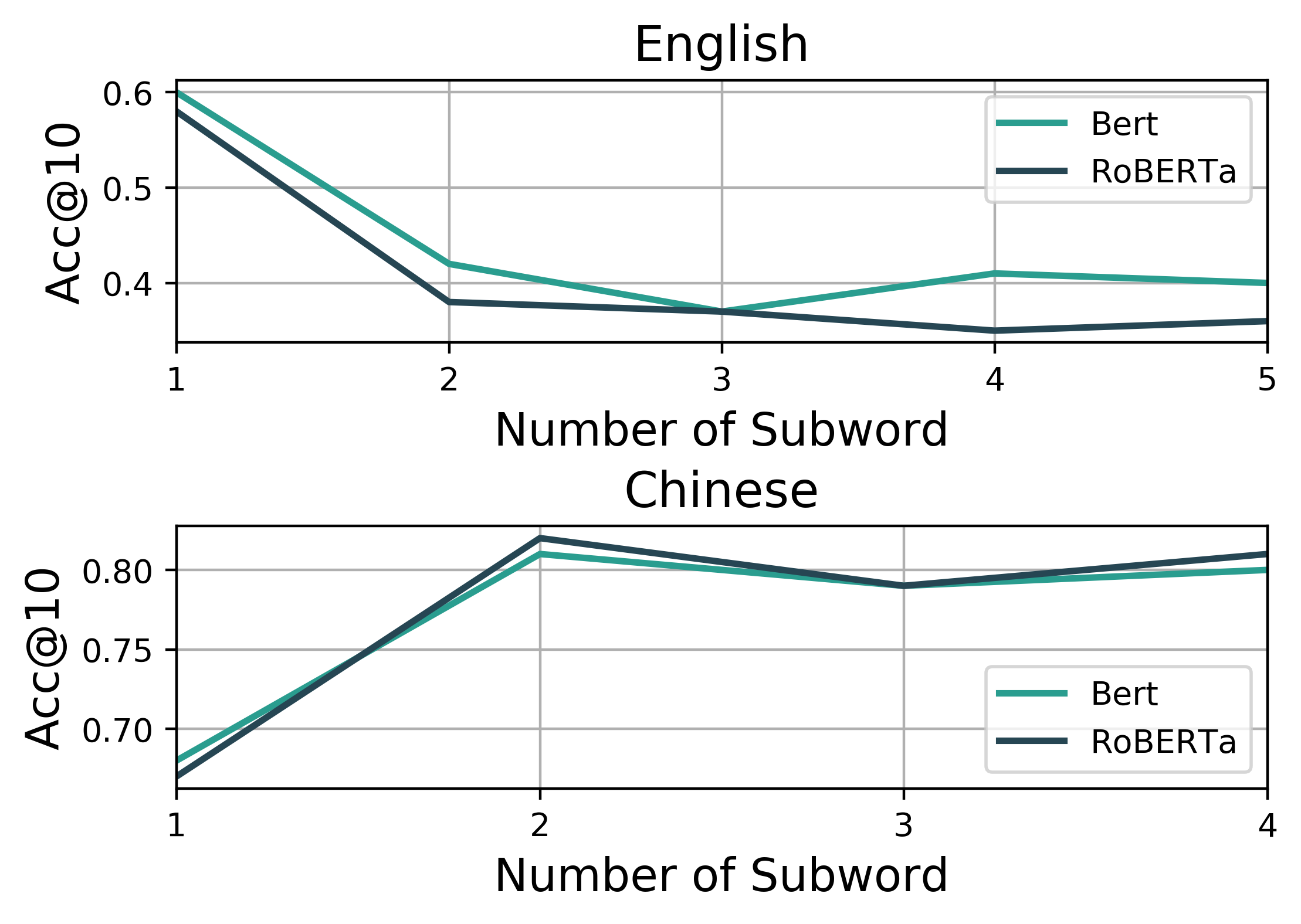}
  \caption{The Acc@10 for words with a different number of subwords. }\label{fig:bpe_lengt_performance}
\end{figure}

Since BERT decomposes words into subwords, we want to investigate whether the number of subwords has an impact on performance. We evaluate the English development set, results are shown in Fig. \ref{fig:bpe_lengt_performance}. The model achieves the best accuracy in English words with one subword and Chinese words with two subwords. This might be caused by the fact that most English words and Chinese words have one subword and two subwords, respectively.

\subsection{Unseen Definition in Unaligned Cross-lingual Reverse Dictionary}
\begin{figure}[h]
    \centering
    \includegraphics[width=0.9\columnwidth]{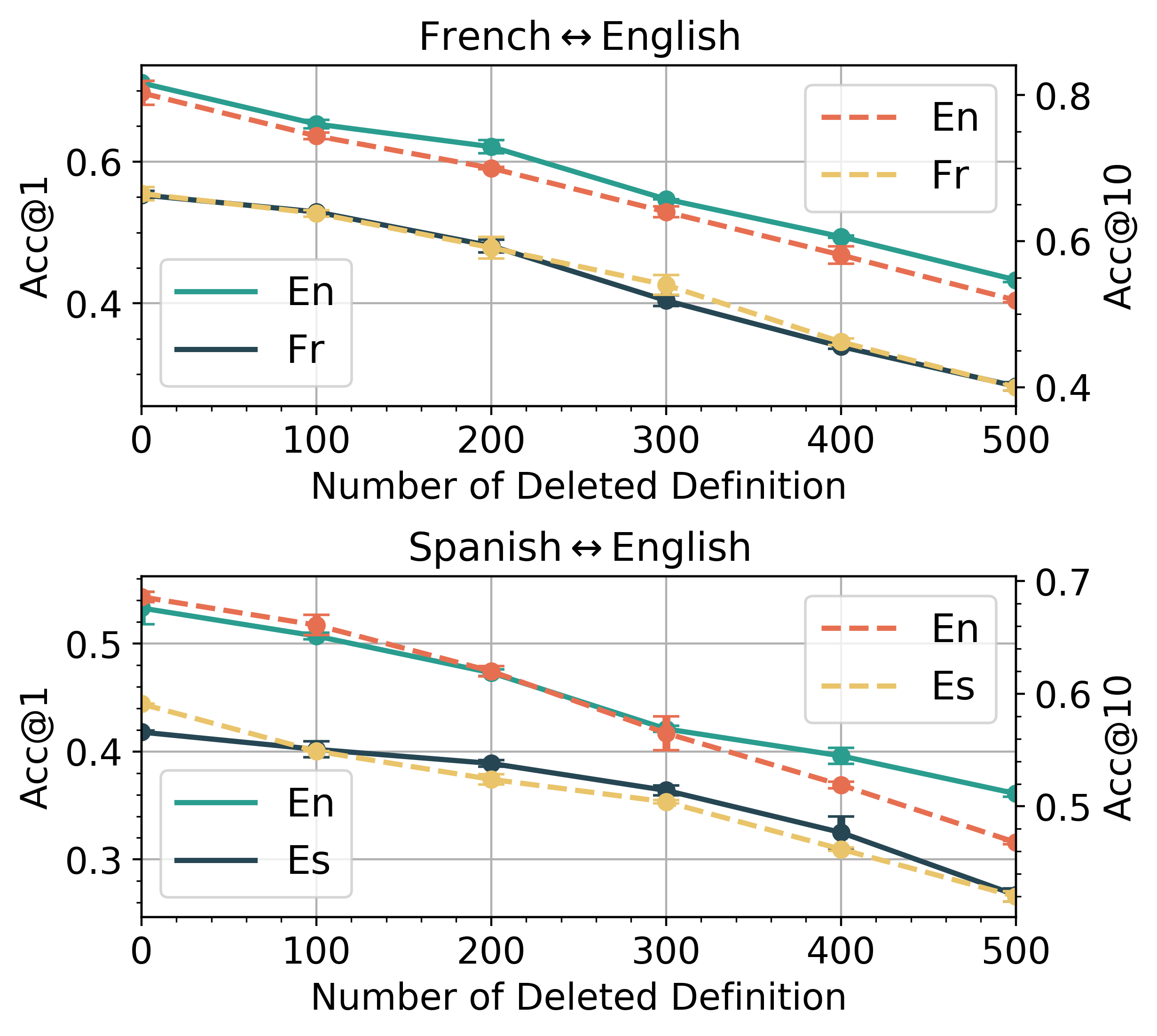}
    \caption{The performance for the unaligned reverse dictionary with the increment of deleted definitions in monolingual data. The dense and dotted lines are Acc@1, Acc@10, respectively. Although the deleted definition and word are in the same language, deleting them harms the performance of cross-lingual word retrieval.}\label{tb:exp5}
\end{figure}

In this section, for the target words presented in bilingual test sets, we gradually remove their definitions from the monolingual training corpus. The performance changing curve is depicted in Table \ref{tb:exp5}. As a reminder, the test sets need to recall target words in another language, while the deleted word and definition are in the same language. Since the number of removed samples is less than 2\% of the monolingual corpus, the performance decay cannot be totally ascribed to the reducing data. Based on Table \ref{tb:exp5}, for the unaligned reverse dictionary task, we can enhance the cross-lingual word retrieval by including more monolingual word definitions.

\begin{CJK}{UTF8}{gbsn}
  \begin{table}[h]
    \small
    \renewcommand\arraystretch{1.1}
    \addtolength{\tabcolsep}{-3.5pt}
    \begin{tabular}{c|l}
    \toprule
    Definition & \begin{tabular}[c]{@{}l@{}}someone who studies secret code systems \\ in order to obtain secret information\end{tabular} \\ \midrule
      Mul-Channel         & cryptographer	cryptologist	spymaster	snoop  \\
      BERT       & \underline{cryptanalyst}  codebreaker  cryptographer coder  \\
      RoBERTa    & codebreaker  \underline{cryptanalyst}  cryptographer  snooper \\ \bottomrule
    \end{tabular}\caption{A Monolingual case displays the advantage of using subwords. In each row is the model's top recalled words; the underlined word is the target word.
    The predicted words by BERT or RoBERTa is either related to ``someone'' (corresponding to the ``-analyst'' or ``er'') or ``code/secret'' (correspoding to ``code-'' or ``crypt-''). 
    }\label{tb:case1}
    \end{table} 
\end{CJK}

\begin{table}[]
  \small
  \addtolength{\tabcolsep}{-3.5pt}
  \begin{tabular}{l|l}
  \toprule
  Definition   & \begin{tabular}[c]{@{}l@{}}El punto que esta a mitad del camino entre dos \\ extremos. (The point that is halfway between \\ two ends)\end{tabular} \\ \midrule
  Spanish           & centro \quad mitad \quad medio \quad punta      \\
    \quad Trans.    & core  \quad middle \quad middle \quad tip  \\
  Unaligned & \underline{center} \quad centre \quad middle \quad mid        \\ \bottomrule
  \toprule
  Definition   & \begin{tabular}[c]{@{}l@{}}Pièce où l ’ on prépare et fait cuire les aliments \\(Room where food is prepared and cooked)\end{tabular} \\ \midrule
  French           & cuisine \quad restaurant \quad pièce \quad cuire      \\
    \quad Trans.    & cookery \quad restaurant \quad room \quad cook  \\
  Unaligned & \underline{kitchen} \quad cook \quad office \quad restaurant        \\ \bottomrule
  \end{tabular} \caption{Unaligned reverse dictionary results by translation and the proposed unaligned reverse dictionary model. The target word is underlined, the ``Trans.'' row is the word translation results. The Spanish ``centro'' in the upper block also has the meaning ``center'', but without context, it gives the wrong translation, and the French word ``cuisine'' in the lower block makes the same error.} \label{tb:case2}
\end{table}

\subsection{Case Study} \label{sec:case_study}
For the monolingual scenario, we present an example in Table \ref{tb:case1} to show that decomposing words into subwords helps to recall related words. Table \ref{tb:case2} shows the comparison between ``mBERT-Trans'' and ``mBERT-joint-Unaligned''.

\section{Conclusion}
In this paper, we formulate the reverse dictionary task under the masked language model framework and use BERT to predict the target word. Since BERT decomposes words into subwords, the score of the target word is the sum of the scores of its constituent subwords. With the incorporation of BERT, our method achieves state-of-the-art performances for both the monolingual and cross-lingual reverse dictionary tasks.
Besides,  we propose a new cross-lingual reverse dictionary task without aligned data. 
Our proposed framework can perform the cross-lingual reverse dictionary while being trained on monolingual corpora only.
Although the performance of unaligned BERT is superior to the translation and word vector alignment method, it still lags behind the supervised aligned reverse dictionary model. Therefore, future work should be conducted to enhance performance on the unaligned reverse dictionary.

\section*{Acknowledgements}
We would like to thank the anonymous reviewers for their insightful comments. We also thank the developers of fastNLP\footnote{\url{https://github.com/fastnlp/fastNLP}}, Yunfan Shao and Yining Zheng, to develop this handy natural language processing package. This work was supported by the National Natural Science Foundation of China (No. 61751201, 62022027 and 61976056), Shanghai Municipal Science and Technology Major Project (No. 2018SHZDZX01) and ZJLab.

\bibliographystyle{acl_natbib}
\bibliography{anthology,emnlp2020}

\end{document}